\pdfoutput=1

\documentclass[11pt]{article}

\usepackage[final]{ACL2023}

\usepackage{times}
\usepackage{latexsym}

\usepackage[T1]{fontenc}

\usepackage[utf8]{inputenc}
\usepackage{graphicx}
\usepackage{microtype}
\usepackage{amsmath}
\usepackage{inconsolata}

\usepackage{colortbl}
\usepackage{xcolor}

\usepackage{multirow}
\usepackage{booktabs}
\usepackage{rotating}

\usepackage{booktabs}
\usepackage{tabularx}
\usepackage{array}

\newcolumntype{Y}{>{\raggedright\arraybackslash}X}

\definecolor{darkred}{RGB}{220,50,50}
\definecolor{lightred}{RGB}{250,180,180}
\definecolor{neutral}{RGB}{255,255,255}
\definecolor{lightgreen}{RGB}{180,230,180}
\definecolor{darkgreen}{RGB}{50,180,50}
\newcommand{\heatcell}[1]{%
  \ifdim #1pt < -60pt \cellcolor{darkred!70}%
  \else\ifdim #1pt < -30pt \cellcolor{lightred!70}%
  \else\ifdim #1pt < 30pt \cellcolor{neutral}%
  \else\ifdim #1pt < 61 pt \cellcolor{lightgreen!70}%
  \else\cellcolor{darkgreen!70}%
  \fi\fi\fi\fi #1}
\title{Attribute-Based Diagnosis of LLM Alignment with Hate Speech Annotations}

\author{
Mohammad Amine Jradi$^{*}$ \qquad Faeze Ghorbanpour$^{*}$ \qquad Alexander Fraser \vspace{.2cm}\\
School of Computation, Information and Technology, TU Munich \\
Munich Center for Machine Learning (MCML)\\
\vspace{.1cm}
{\tt \small \{amine.jradi, faeze.ghorbanpour\}@tum.de}\\
\vspace{.1cm}
{\small $^{*}$Equal contribution.}
}

\begin{document}
\maketitle
\begin{abstract}
Hate speech annotation is costly, subjective, and prone to annotator disagreement, making large-scale dataset construction challenging. We systematically analyze how well large language models (LLMs) align with human judgments across ten theoretically grounded subjective attributes, such as dehumanization, violence, and sentiment, evaluating both small and large variants of Llama 3.1 and Qwen 2.5. Our analysis reveals a consistent split across all models: behaviorally explicit dimensions (insult, humiliate, attack-defend) correlate strongly with human annotations, while evaluative dimensions (respect, sentiment, hate speech) are systematically inverted. Demographic persona conditioning reduces model confidence without improving alignment. 
Building on these insights, we propose combining attribute-level LLM predictions via a confidence-weighted Ridge regression to reconstruct continuous hate speech scores from the Measuring Hate Speech corpus, achieving $R^2$ of up to 0.71 and outperforming direct prompting baselines, demonstrating that structured attribute decomposition recovers a richer and more human-aligned signal than end-to-end label prediction alone.
\end{abstract}

\section{Introduction}


Hate speech annotation is a challenging task; determining whether a comment is hateful requires nuanced judgment, as what one person perceives as harmful, another may view as acceptable~\cite{davani-etal-2022-dealing, sap-etal-2022-annotators}. 
This subjectivity manifests as systematic annotator disagreement that cannot be resolved by adjudication alone, and collecting high-quality annotations at scale remains prohibitively expensive and methodologically complex \cite{waseem-hovy-2016-hateful}.
These challenges have motivated a growing line of work on using LLMs as scalable substitutes for human annotators \cite{Argyle_2023, gilardi2023chatgpt, tan2024large}. 
By prompting models to produce labels, researchers have shown promising alignment with human judgments on various tasks \citep{tan2024large}, specifically for subjective tasks~\cite{hu-collier-2024-quantifying, deshpande-etal-2023-toxicity}. 
However, existing work treats annotation simulation as a single holistic prediction, a label or scalar score, offering no insight into which aspects of a subjective construct the model understands and where it diverges from human perception.

Hate speech is not a monolithic concept but a constellation of distinct behaviors and dimensions. The Measuring Hate Speech corpus \cite{kennedy2020constructing} captures this complexity through ten theoretically grounded annotation attributes, including dehumanization, violence, respect, etc., each targeting a specific facet of hateful content. These attributes are combined via Item Response Theory (IRT) into a continuous hate speech score, 
which makes it a unique lens for studying LLM annotation simulation at the attribute level, rather than collapsing hate speech into a single judgment. 

In this paper, we exploit this structure to conduct a systematic investigation of LLM-based subjective tasks annotation prediction. We prompt small and large variants of Llama 3.1 and Qwen 2.5 to rate each of the ten attributes independently, extracting both predictions and model confidence scores. This fine-grained view reveals a consistent and interpretable pattern: behaviorally explicit dimensions such as insult, humiliate, and attack\_defend align strongly with human judgments across four models, while evaluative dimensions such as respect, sentiment, and hate speech itself are systematically inverted, a finding that holds regardless of model size or family. We also examine whether demographic persona conditioning improves this alignment, finding that while it reduces model confidence, it does not bring predictions closer to human labels~\citep{hu-collier-2024-quantifying, sarumi2025impact}. 

Building on these insights, we ask whether fine-grained attribute predictions can be combined to recover the continuous IRT hate speech score more effectively than direct prompting. We propose a confidence-weighted Ridge regression pipeline that aggregates attribute-level LLM predictions into a reconstructed hate speech score, and evaluate it against direct prompting baselines. Our pipeline achieves $R^2$ of up to 0.71 and outperforms direct prompting, demonstrating that structured attribute decomposition captures a richer and more human-aligned signal than end-to-end label prediction.

\section{Related Work}


A growing body of recent studies explores using LLMs to replace or augment human annotators, particularly for subjective tasks where collecting labels at scale is expensive. \citet{calderon2025alternativeannotatortestllmasajudge} propose a statistical framework for testing when LLMs can substitute human annotators, while \citet{radharapu-etal-2025-arbiters} show that LLMs struggle in tasks without annotator consensus. 
\citet{falk-lapesa-2025-mining} examine the relationship between human label variation and model uncertainty in perspectivist annotation, and \citet{chochlakis-etal-2025-aggregation} warn that naive aggregation of LLM outputs can collapse meaningful variation in subjective tasks. Together, these works establish that LLM annotation simulation is promising but fundamentally limited, limitations that our work examines at the attribute level rather than the label level. 


Hate speech annotation is particularly susceptible to subjectivity and cultural variation. \citet{lee-etal-2024-exploring-cross} show that hate speech annotations vary systematically across cultures, motivating perspectivist approaches to modeling. \citet{alacam-etal-2025-disentangling} distinguish subjectivity from uncertainty in hate speech modeling, while \citet{lu2025llm} analyze LLM confidence under offensive language disagreement. 
\citet{giorgi2025human} compare human and persona-based LLM biases in hate speech annotation, finding that LLM biases differ systematically from human annotator biases, a direction our work extends by identifying which specific attributes align, which invert, and how this pattern holds across models. \citet{gligoric-etal-2025-unconfident} use LLM confidence to decide when human annotation is needed; we instead treat confidence as a weighting signal for attribute-based score prediction. 


\section{Our Approach}


We use LLMs to rate each of the ten MHS attributes independently for each comment, under two conditions: a \textit{vanilla} prompt with no demographic context and a \textit{persona} prompt conditioned on the real demographic profile of the corresponding MHS annotator. For each prediction, we extract a confidence score from token-level log probabilities and measure attribute-level alignment with human judgments via Spearman rank correlation. 
Building on these predictions, we construct confidence-weighted attribute features, which are passed to a Ridge regression model trained on the training set to reconstruct the continuous IRT hate speech score on the test set.

\begin{table*}[t]
\centering
\small
\begin{tabular}{p{1cm}p{1cm}p{0.9cm}p{0.9cm}p{0.9cm}p{0.9cm}p{0.9cm}p{1.1cm}p{0.9cm}p{0.9cm}p{0.9cm}p{0.5cm}}
\toprule
& & {Respect} & {Sentiment} 
& {Status} & {Hate speech}& {Genocide} 
& {Dehuman- ize} & {Attack- defend} & {Violence} & {Humiliate} & {Insult} \\
\midrule
\multirow{2}{*}{Llama 70B}
& Vanilla & \heatcell{-73.03} & \heatcell{-70.47} 
& \heatcell{-57.35} & \heatcell{-48.79} & \heatcell{54.87} & \heatcell{59.78} & \heatcell{61.76} & \heatcell{64.45} & \heatcell{66.42} & \heatcell{69.75} \\
& Persona & \heatcell{-75.67} & \heatcell{-72.40} 
& \heatcell{-58.21} & \heatcell{-50.01} & \heatcell{53.79} & \heatcell{60.27} & \heatcell{60.99} & \heatcell{68.68} & \heatcell{67.43} & \heatcell{71.65} \\
\midrule
\multirow{2}{*}{{Qwen 72B}} 
& Vanilla & \heatcell{-74.44} & \heatcell{-72.78} 
& \heatcell{-56.11} & \heatcell{-51.34} & \heatcell{52.60} & \heatcell{57.87} & \heatcell{61.17} & \heatcell{64.81} & \heatcell{66.70} & \heatcell{71.17} \\
& Persona & \heatcell{-74.19} & \heatcell{-71.74} 
& \heatcell{-56.03} & \heatcell{-52.27} & \heatcell{53.31} & \heatcell{57.77} & \heatcell{60.82} & \heatcell{66.83} & \heatcell{66.69} & \heatcell{70.68} \\
\bottomrule
\end{tabular}
\caption{Spearman rank correlation ($\times$100) between LLM attribute predictions and human ratings for large models. 
}
\label{tab:spearman}
\end{table*}

\noindent
\paragraph{Attribute-level simulation:}
\label{sec:attribute-simulation}
Let $\mathcal{A} = \{a_1, \dots, a_{10}\}$ denote the ten MHS attributes and $\mathcal{F} = \{f_1,  \dots, f_m\}$ denote the $m$ demographic features of each annotator (gender, age, race, religion, and  political ideology). For each comment $n$ and attribute $a_i$, we prompt the LLM to predict an 
ordinal label $S_{n,i} \in \{0, \dots, s\}$ \footnote{$s$ represent the maximum ordinal value that varies for each attribute.} under two conditions:
In the \textit{vanilla} condition, the model  receives only the attribute definition and comment text with no demographic context:
$
S_{n,i} = \text{LLM}(a_i, c_n)$. In the \textit{persona} condition, the prompt is additionally conditioned on the annotator's demographic profile $\mathbf{f}_n \in \mathcal{F}$:
$ S_{n,i} = \text{LLM}(a_i, c_n, \mathbf{f}_n)$.

\noindent
\paragraph{Confidence extraction:} To evaluate attribute-level alignment, we compute the Spearman rank correlation between LLM predictions $S_{n,i}$ and mean human ratings ${h}_{n,i}$ for each attribute $a_i$:
$\rho_i = \mathrm{Spearman}(S_{n,i}, {h}_{n,i})$,
Confidence scores are extracted from token-level log probabilities, renormalized via softmax $\mathcal{V}$:
$
C_{n,i} = \frac{\exp(\ell_{\hat{k}})}
{\sum_{k \in \mathcal{V}} \exp(\ell_k)},
$
where $\ell_k$ is the log probability of label token $k$ and $\hat{k} = \arg\max_{k \in \mathcal{V}} P(k \mid \text{prompt})$.

\noindent
\paragraph{Score reconstruction:}
\label{sec:score-reconstruction}
The confidence--weighted feature for each attribute is: $x_{n,i} = S_{n,i} \cdot C_{n,i}$, and the final hate speech score is predicted by a Ridge regression model trained on the training set:
$\hat{y}_n = \text{Ridge}(\mathbf{x}_n)$, where $\mathbf{x}_n = (x_{n,1}, \dots, x_{n,10})$ is the vector of weighted attribute features. 




\section{Experimental Details}

\paragraph{Dataset}


We use the Measuring Hate Speech (MHS) corpus \cite{kennedy2020constructing}, a large-scale dataset of 39,565 comments annotated by 7,912 crowd-sourced annotators. 
MHS captures the multidimensional nature of hateful content through ten ordinal attributes, each targeting a distinct facet: \textit{hatespeech} (whether the comment contains hate speech), \textit{attack/defend} (whether it attacks or defends a group), \textit{genocide} (whether it calls for or justifies genocide), \textit{violence} (whether it incites or glorifies violence), \textit{dehumanize} (whether it dehumanizes a group), \textit{status} (whether it degrades the status of a group), \textit{humiliate} (whether it humiliates a group or individual), \textit{insult} (whether it contains insults), \textit{respect} (whether it shows respect toward a group), and \textit{sentiment} (the overall sentiment toward the target group). 
The final hate speech score is derived via Item Response Theory (IRT), a psychometric framework that aggregates attribute ratings into a continuous score while accounting for annotator reliability \citep{sachdeva2022assessing}. 

\paragraph{Models}

We evaluate four open-source LLMs across two model families and two size regimes: Meta-Llama-3.1-70B-Instruct and Meta-Llama-3.1-8B-Instruct \cite{grattafiori2024llama}, and Qwen2.5-72B-Instruct and Qwen2.5-7B-Instruct \cite{yang2025qwen2.5}. 
Evaluating models of different sizes and families allows us to assess whether alignment patterns generalize beyond a single model or scale. 
Details on the models are in Appendix~\ref{sec:models}.

\paragraph{Prompts}

We design two prompt variants to simulate attribute annotation. The \textit{vanilla} prompt presents the model with the attribute definition, its ordinal scale anchors drawn from the original MHS annotation guidelines \cite{sachdeva2022measuring}, and the comment text. Grounding the prompts in the original guidelines ensures a fair and consistent comparison between LLM and human judgments across all ten attributes.
The \textit{persona} prompt extends the vanilla prompt by prepending the real demographic profile of the corresponding MHS annotator. 
For the baseline comparison, we design four direct binary prompting variants: a zero-shot prompt with no additional context, a few-shot prompt with example annotations, a definition-based prompt providing a brief description of hate speech, and an attribute-aware prompt listing all ten MHS attributes as considerations without decomposing them. 
All four ask the model to output a single hate/non-hate label and are compared against our attribute-level pipeline on binary classification performance. Additional details about the prompts can be found in Appendix \ref{sec:prompts}.

\section{Results}
We structure our results in two parts. First, we analyze attribute-level LLM-human alignment across the full dataset using Spearman's rank correlation. Second, we evaluate hate speech score reconstruction using 5-fold cross-validation on the train/test split, reporting R², F1, accuracy, precision, and recall averaged across folds. 

\paragraph{Attribute-level Analysis}


Table~\ref{tab:spearman} presents the Spearman rank correlations between LLM attribute predictions and human ratings across all four models and prompting conditions. A consistent pattern emerges across all settings: attributes naturally split into two clusters. Behaviorally explicit attributes, such as insult, humiliate, violence, attack\_defend, dehumanize, and genocide, show strong positive correlations with human judgments (ranging from 53 to 72), indicating that LLMs reliably track human perception of concrete, observable behaviors in text. In contrast, evaluative attributes, such as respect, sentiment, status, and hate speech, are systematically inverted (ranging from -52 to -75), suggesting that LLMs apply a different judgment schema for dimensions that require holistic or contextual assessment. Results for smaller models are in Appendix~\ref{sec:small_models}.

Figure~\ref{fig:confidence} shows the relationship between average model confidence and Spearman correlation with human ratings for each attribute, across both larger models and prompting conditions. A notable pattern emerges: the attributes that align most poorly with human judgments, respect, sentiment, and hate speech cluster in the bottom-right quadrant, indicating that the model is highly confident precisely where it diverges most from human perception. While positively correlated attributes such as insult, humiliate, and violence also attract relatively high confidence, the negatively correlated attributes tend to sit at the highest confidence range (0.88–0.95), suggesting that LLMs are not only wrong on evaluative attributes but particularly certain about their wrong judgments. This pattern holds consistently across both large models. In smaller models, the confidence range is lower, and the two-cluster separation is less pronounced; confidence plots for smaller models are in Appendix~\ref{sec:small_models}. Persona conditioning consistently shifts attribute confidence leftward across both large models without improving alignment, confirming that demographic conditioning affects model certainty but not its conceptual understanding of attributes.

\begin{figure}[t]
\centering
\includegraphics[width=\linewidth]{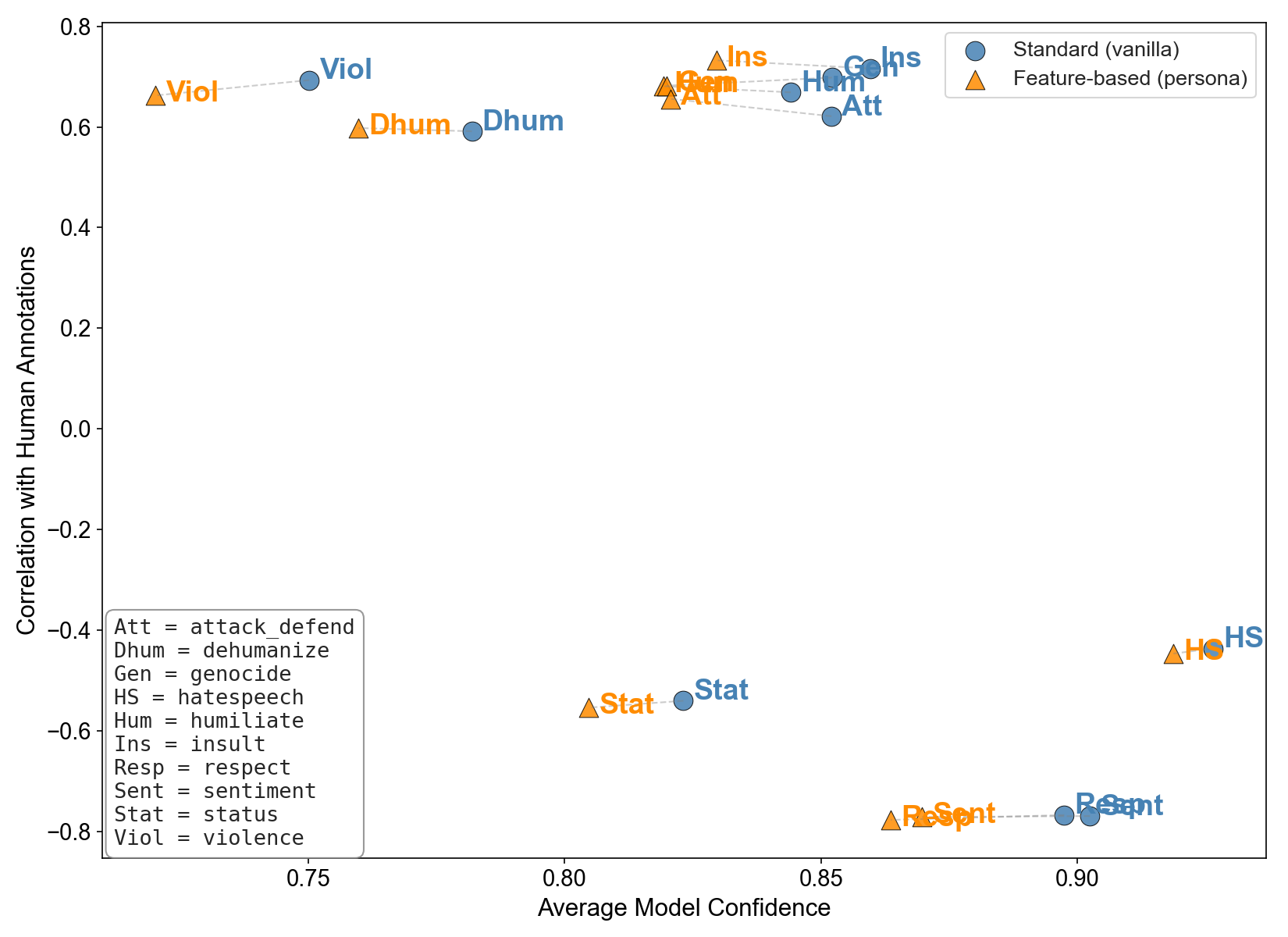}
\includegraphics[width=\linewidth]{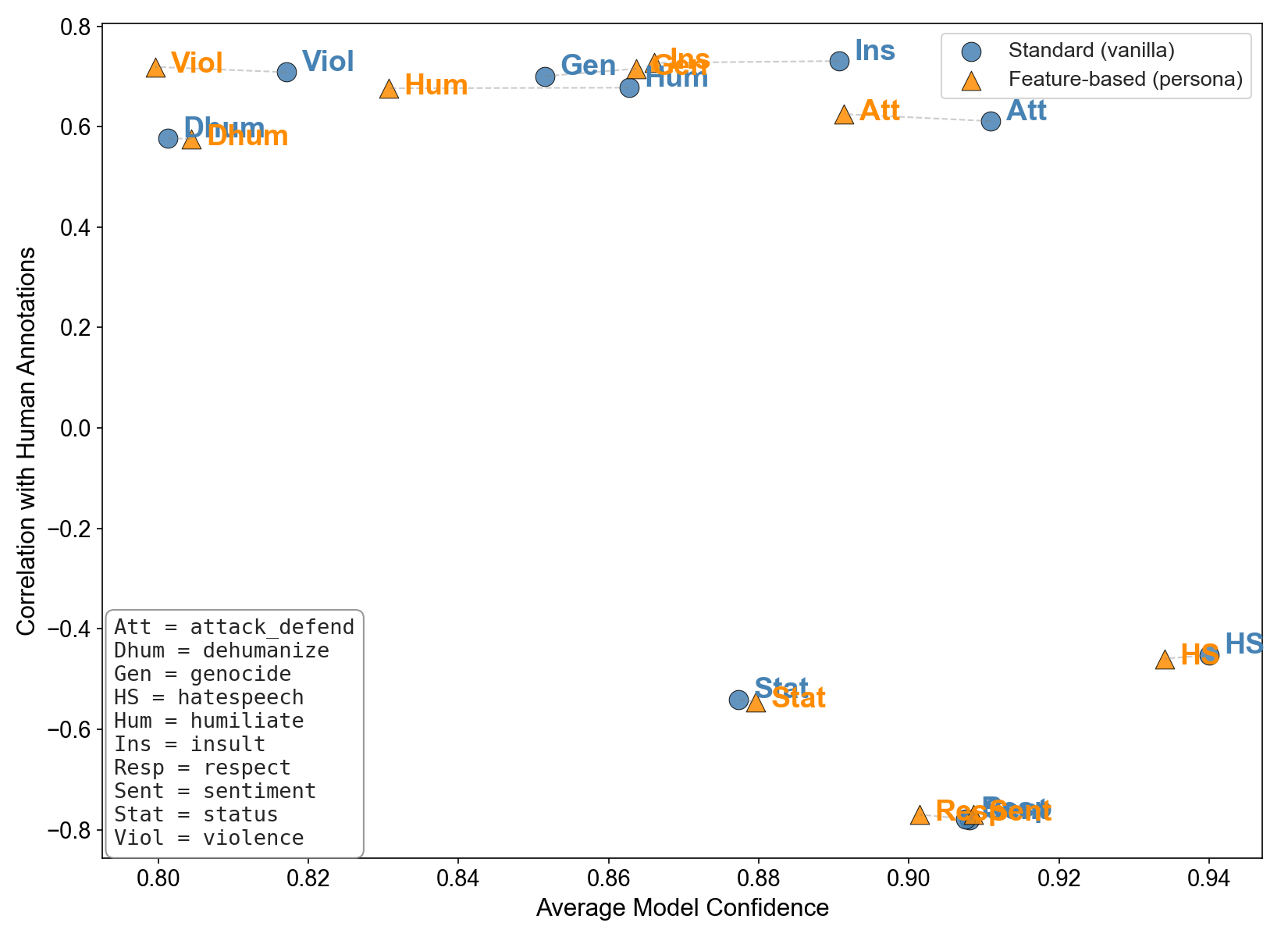}
\caption{Average model confidence versus Spearman correlation with human ratings for each attribute.
}
\label{fig:confidence}
\end{figure}


\paragraph{Hate Score Reconstruction}


Table~\ref{tab:results} presents the hate speech score reconstruction and classification results across larger models, prompting conditions, and baselines. Our confidence-weighted Ridge pipeline achieves R² values of up to 70.70 for Llama-3.1 and 68.91 for Qwen-2.5, demonstrating that attribute-level LLM predictions can reliably reconstruct the continuous IRT hate speech score without access to ground-truth labels at test time.

Compared to direct prompting baselines, our pipeline consistently achieves superior accuracy (~84\% vs 57--70\%) and competitive F1 across both models. While Qwen baselines show competitive F1, this masks extreme prediction bias — baselines achieve very high recall (up to 99\%) but poor precision (as low as 34\%), flagging the vast majority of comments as hate speech regardless of content. Our pipeline produces substantially more calibrated predictions, achieving a better balance between precision and recall and consequently higher accuracy and F1. Llama-3.1 outperforms Qwen-2.5 despite Qwen showing marginally stronger Spearman correlations on individual attributes in Table~\ref{tab:spearman}, suggesting that aggregate reconstruction quality does not directly follow from per-attribute alignment. 
See Appendices~\ref{sec:ablation} and \ref{sec:examples} for the ablation study and qualitative examples of model predictions.






\begin{table}[t]
\centering
\small
\begin{tabular}{p{0.1cm}p{1.4cm}p{1.5cm}cccc}
\toprule
& & {R²} & {F1} & {Acc.} & {Prec.} & {Recall} \\
\midrule
\multirow{7}{*}{\rotatebox[origin=c]{90}{Llama-70B}}
& Zero-shot & -- & 60.30 & 66.64 & 43.95 & 96.02 \\
& Few-shot & -- & 60.63 & 67.47 & 44.53 & 94.93 \\
& Definition & -- & 61.61 & 70.79 & 47.15 & 88.88 \\
& Attr. aware & -- & 60.30 & 66.97 & 44.15 & 95.12 \\
& Attr. value & -- & 54.91 & 57.49 & 38.12 & 98.14 \\
& Vanilla & 70.57 {\scriptsize $\pm$ 0.59} & 68.63 & 84.02 & 71.15 & 66.29 \\
& Persona & 70.71 {\scriptsize $\pm$ 0.45} & 68.95 & 84.49 & 73.07 & 65.27 \\
\midrule
\multirow{7}{*}{\rotatebox[origin=c]{90}{Qwen-72B}}
& Zero-shot & -- & 70.60 & 70.03 & 54.86 & 99.02 \\
& Few-shot & -- & 50.33 & 51.36 & 34.62 & 99.40 \\
& Definition & -- & 69.64 & 69.11 & 54.17 & 97.49 \\
& Attr. aware & -- & 69.45 & 68.26 & 53.42 & 99.26 \\
& Attr. value & -- & 50.71 & 48.98 & 34.02 & 99.48 \\
& Vanilla & 68.85 {\scriptsize $\pm$ 0.56} & 69.20 & 83.95 & 70.08 & 68.34 \\
& Persona & 68.65 {\scriptsize $\pm$ 0.60} & 67.63 & 83.83 & 71.67 & 64.03 \\
\bottomrule
\end{tabular}
\caption{Hate speech score reconstruction and 
classification results ($\times$100). 
}
\label{tab:results}
\end{table}

\section{Conclusion}
We investigated LLM alignment with hate speech annotations at the attribute level, revealing a consistent split across four models: behaviorally explicit attributes align well with human judgments, while evaluative attributes are systematically inverted. 
Our confidence-weighted Ridge pipeline achieves $R^2$ of up to 0.71 on hate speech score prediction, demonstrating that structured attribute decomposition recovers a stronger signal than direct label prediction. 

\section*{Limitations}
Our study focuses on the Measuring Hate Speech corpus, as it is the only dataset providing the combination of multi-attribute annotation, continuous IRT scoring, and annotator demographic profiles required by our pipeline; extending to other corpora would require comparable annotation depth. We evaluate two model families, Llama and Qwen, in two size regimes; broader coverage was constrained by computational resources, though the consistency of findings across four models suggests the patterns are not family-specific. Persona conditioning relies on demographic features available in MHS; richer annotator signals, such as annotation history or cultural context, are not provided in the dataset and cannot be incorporated. 

\section*{Ethical Considerations}
This work involves processing hateful content, which may pose risks to researchers and annotators involved in data collection and model evaluation. We use an existing dataset and do not collect new annotations. While our pipeline could reduce the cost of hate speech annotation, we caution against fully replacing human annotators, particularly given the systematic misalignments identified on evaluative attributes, dimensions that often capture the most nuanced and culturally grounded aspects of hate speech. Automated annotation systems should be used to assist rather than substitute human judgment in sensitive content moderation.

\bibliography{custom}
\bibliographystyle{acl_natbib}

\appendix

\nocite{ghorbanpour-etal-2025-fine, ghorbanpour-etal-2025-data}

\section{Models}
\label{sec:models}
All models used in this study are loaded in quantized form and served via vLLM \cite{kwon2023efficient} from HuggingFace \cite{wolf-etal-2020-transformers}. 
We load them in inference mode, without training and changing their weights. 
To ensure deterministic and comparable outputs across runs, we set the temperature to 0.0, fix the random seed to 42, and limit generation to a single token (max\_tokens=1), enabling direct extraction of token-level probabilities as confidence scores for each attribute prediction.


For the {Spearman correlation analysis} 
(Section~\ref{sec:attribute-simulation}), we retain 
the full (comment, annotator) granularity. Each LLM 
prediction is compared against the corresponding 
human annotator's label, allowing a direct 
per-annotator alignment measurement.
For the {Ridge regression} 
(Section~\ref{sec:score-reconstruction}), we operate 
at the comment level. The IRT-derived hate speech 
score $\theta_n$ is computed once per comment, so 
predictions must also be aggregated to the comment 
level. Under the vanilla condition, this is 
straightforward. Under the persona condition, the 
LLM produces one prediction per (comment, annotator) 
pair — these are averaged across annotators before 
entering Ridge, ensuring a consistent unit of 
analysis across both conditions. After merging on 
\texttt{comment\_id}, the regression set comprises 
approximately 39,500 unique comments.

\begin{figure}[t]
\centering
\includegraphics[width=\linewidth]{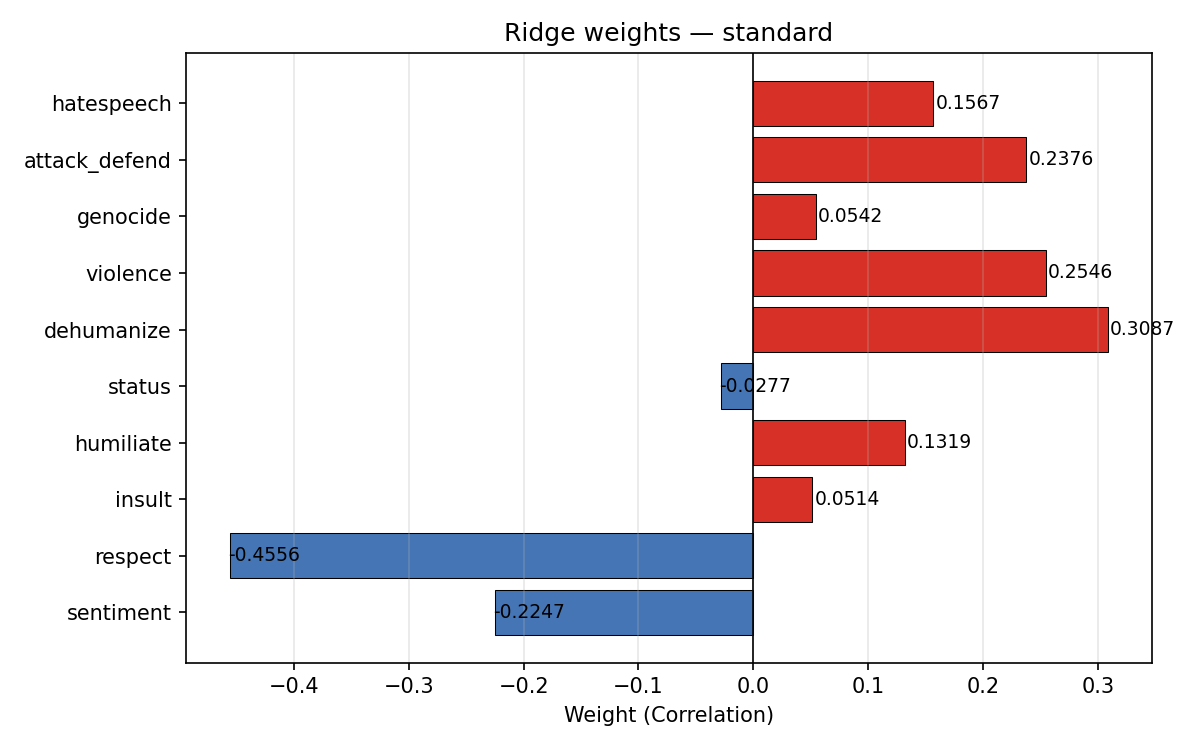}\\[2pt]
\includegraphics[width=\linewidth]{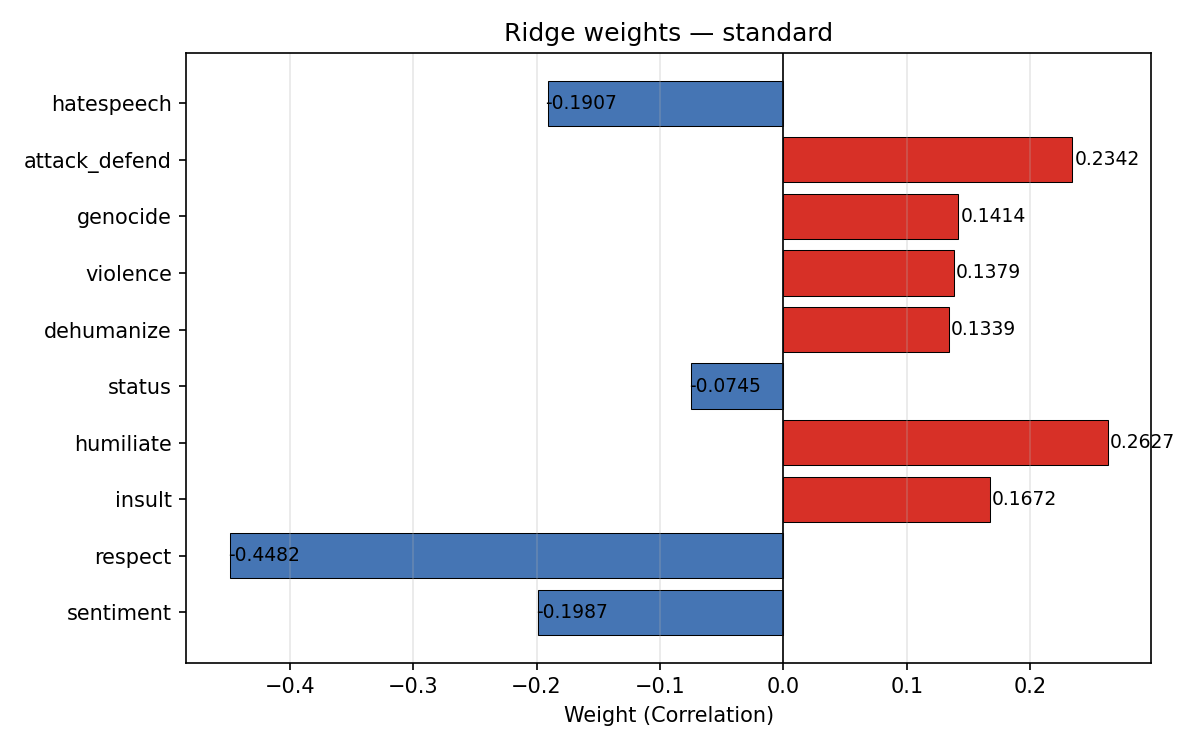}
\caption{Ridge regression weights per attribute 
for two models (LLama top and Qwen bottom). Positive weights 
indicate hate-promoting attributes, negative 
weights reflect evaluative attributes inverted 
relative to human judgments (cf. Table~\ref{tab:spearman}).}
\label{fig:llama_weights}
\end{figure}

Figure~\ref{fig:llama_weights} shows the Ridge weights for each attribute across vanilla conditions for both models. Attributes with strong positive Spearman correlations in Table~\ref{tab:spearman}, insult, humiliate, violence, receive large positive weights, while evaluative attributes, respect, sentiment, hate speech, receive large negative weights. This direct correspondence between the alignment analysis in section 1 and the learned weights confirms that the Ridge model exploits the inversion signal from evaluative attributes rather than discarding it, turning a potential weakness into a useful predictive feature.

\section{Prompts}
\label{sec:prompts}
All prompts share a common structure consisting of a system prompt instructing the model to act as an expert annotator without applying safety filters, followed by a user prompt containing the attribute description with ordinal scale anchors drawn from the original MHS annotation guidelines \cite{sachdeva2022measuring}, the comment text, and an instruction to output a single integer token. The per-attribute scoring rubrics used in the vanilla condition are shown in Table~\ref{tab:prompts-by-attribute}.
In the vanilla condition, the system prompt provides no demographic context. In the persona condition, a demographic header (Table~\ref{tab:persona-header}) is prepended to each per-attribute rubric. The header takes the form: \textit{"You are a content annotator with the following characteristics: gender ..., age ... (old/young), race ..., religion ..., ideology .... Evaluate this comment from your perspective as this annotator."} Demographic fields are populated from the corresponding row of the MHS annotation file, so the model produces one prediction per (comment, annotator, attribute) triple, mirroring the granularity of the human annotation data. The rubric itself is unchanged between the two conditions, and both use identical decoding configurations, ensuring that the only difference between vanilla and persona predictions is the demographic context.

\begin{table*}[ht]
    \centering
    \begin{tabular}{@{}p{0.92\textwidth}@{}}
        \toprule
        \textbf{Persona header} \\
        \midrule
        \texttt{You are a content annotator with the following characteristics:} \\
        \texttt{- Gender: \{gender\}} \\
        \texttt{- Age: \{age\} (\{age\_category\})} \\
        \texttt{- Race: \{race\}} \\
        \texttt{- Religion: \{religion\}} \\
        \texttt{- Ideology: \{ideology\}} \\
        \texttt{Evaluate this comment from your perspective as this annotator.} \\
        \bottomrule
    \end{tabular}
    \caption{Persona header prepended to each per-attribute scoring rubric (Table~\ref{tab:prompts-by-attribute}) in the persona prompting condition. Placeholders in curly braces are populated per row from the demographic fields of the corresponding MHS annotator.}
    \label{tab:persona-header}
\end{table*}

\begin{table*}[t]
\centering
\small
\setlength{\tabcolsep}{4pt}
\renewcommand{\arraystretch}{1.12}
\begin{tabularx}{\textwidth}{@{}p{0.14\textwidth} p{0.08\textwidth} Y@{}}
\toprule
\textbf{Attribute} & \textbf{Scale} & \textbf{Scoring rubric} \\
\midrule

\texttt{sentiment} & 0--4 &
Sentiment polarity: 0=strongly negative, 1=somewhat negative, 2=neutral, 3=somewhat positive, 4=strongly positive. \\

\texttt{hatespeech} & 0--2 &
Presence of hate speech: 0=no, 1=unclear/neutral, 2=yes. \\

\texttt{insult} & 0--4 &
Insult toward the group: 0=none, 1=mild, 2=neutral/unsure, 3=clear, 4=severe. \\

\texttt{humiliate} & 0--4 &
Humiliation toward the group: 0=none, 1=mild, 2=neutral/unsure, 3=attempted humiliation, 4=degrading. \\

\texttt{dehumanize} & 0--4 &
Dehumanization of the group: 0=strongly no, 1=no, 2=unclear/neutral, 3=yes, 4=strongly yes. \\

\texttt{violence} & 0--4 &
Call for violence against the group: 0=strongly no, 1=no, 2=unclear/neutral, 3=yes, 4=strongly yes. \\

\texttt{genocide} & 0--4 &
Call for deliberate large-scale killing of the group: 0=strongly no, 1=no, 2=unclear/neutral, 3=yes, 4=strongly yes. \\

\texttt{status} & 0--4 &
Relative social status framing: 0=strongly inferior, 1=inferior, 2=equal/neutral, 3=superior, 4=strongly superior. \\

\texttt{respect} & 0--4 &
Respect toward the group: 0=strongly disrespectful, 1=disrespectful/rude, 2=neutral, 3=respectful/polite, 4=strongly respectful. \\

\texttt{attack\_defend} & 0--4 &
Stance toward the group: 0=strongly defending, 1=defending, 2=neutral/mixed, 3=attacking, 4=strongly attacking. \\

\bottomrule
\end{tabularx}
\caption{Per-attribute scoring rubrics used in the vanilla prompting condition. Each attribute is queried with a fixed prompt defining the ordinal scale. In the persona condition, the persona header in Table~\ref{tab:persona-header} is prepended verbatim; the rubric itself is unchanged.}
\label{tab:prompts-by-attribute}
\end{table*}

\section{Small Model Results}
\label{sec:small_models}
\begin{figure}[t]
\centering
\includegraphics[width=\linewidth]{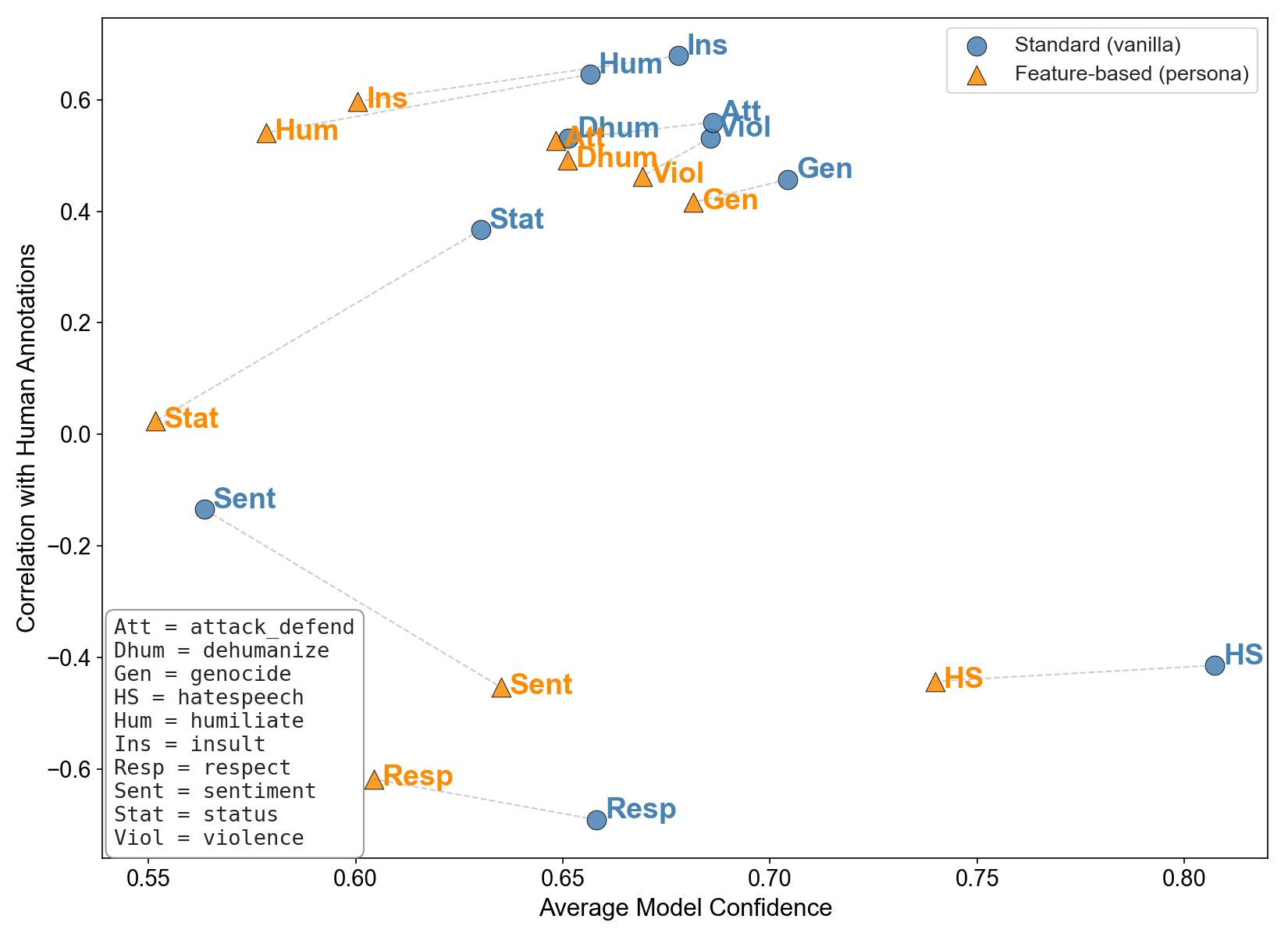}
\includegraphics[width=\linewidth]{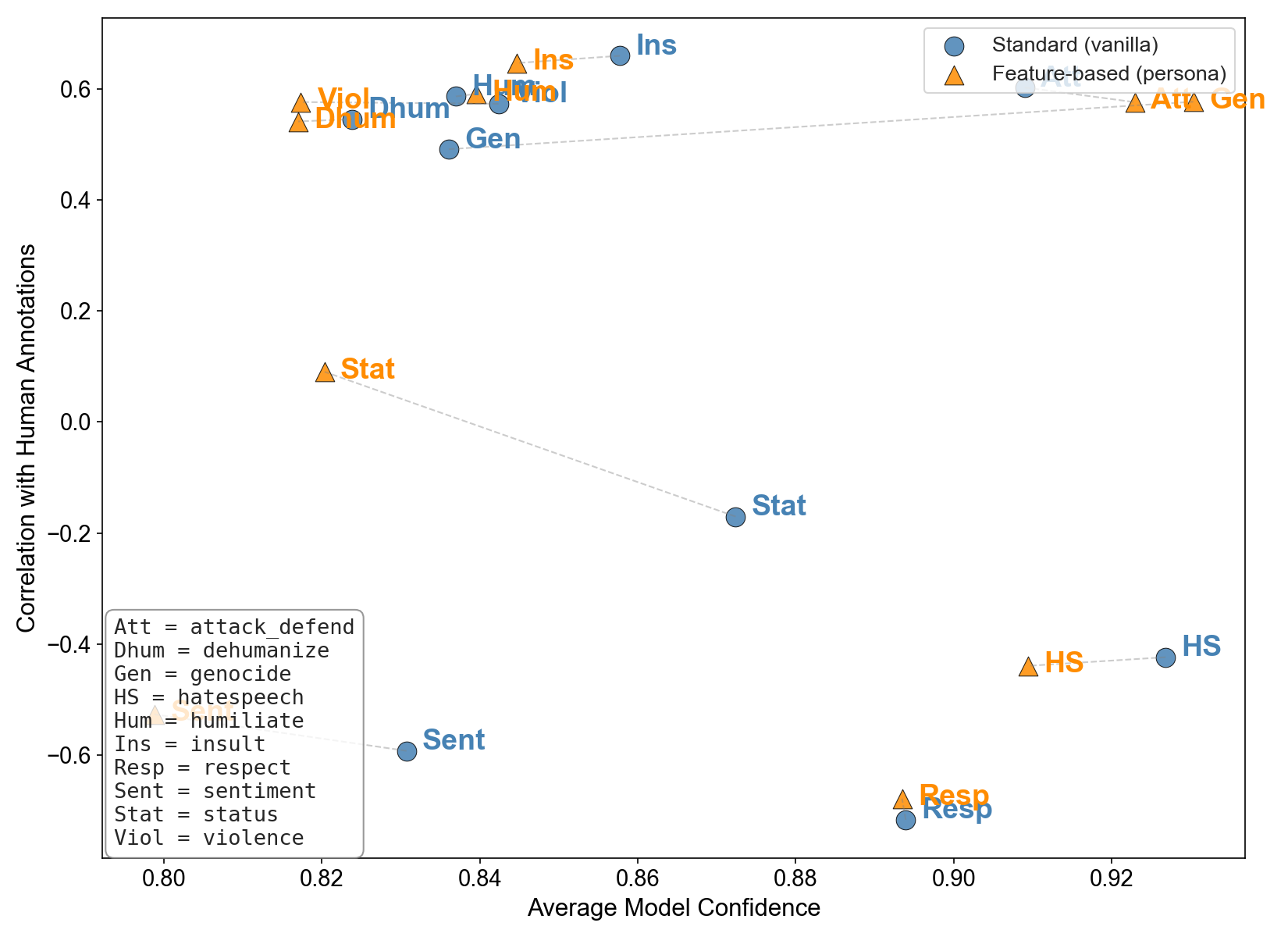}
\caption{Average model confidence versus Spearman correlation with human ratings for each attribute.
}
\label{fig:confidence-small}
\end{figure}

Table~\ref{tab:spearman-small} presents the Spearman rank correlations for the two smaller models. The two-cluster pattern observed in large models is largely preserved, though with notable deviations. In Llama-3.1-8B, sentiment collapses to near zero (-1.56) under vanilla prompting and status flips to positive (+38.68), suggesting that smaller models struggle to maintain consistent evaluative judgments on these dimensions. Qwen-2.5-7B shows a more stable pattern, retaining negative correlations on all evaluative attributes, though with weaker magnitudes than its 72B counterpart.
Figure~\ref{fig:confidence-small} shows the confidence vs. correlation plots for both small models. Unlike large models where evaluative attributes cluster clearly in the bottom-right quadrant, small models show a more dispersed pattern with lower overall confidence (0.55--0.80), and the separation between behaviorally explicit and evaluative attributes is less pronounced.

\begin{table*}[t]
\centering
\small
\begin{tabular}{p{1cm}p{1cm}p{0.9cm}p{0.9cm}p{0.9cm}p{0.9cm}p{0.9cm}p{1.1cm}p{0.9cm}p{0.9cm}p{0.9cm}p{0.5cm}}
\toprule
& & {Respect} & {Sentim- ent} & {Status} & {Hate speech} & {Genoc- ide} & {Dehuma- nize} & {Attack-defend} & {Violen- ce} & {Humili- ate} & {Insult} \\
\midrule
\multirow{2}{*}{Llama-8B}
& Vanilla & \heatcell{-70.25} & \heatcell{-1.56} & \heatcell{38.68} & \heatcell{-46.69} & \heatcell{43.03} & \heatcell{55.40} & \heatcell{58.66} & \heatcell{54.99} & \heatcell{64.79} & \heatcell{67.47} \\
& Persona & \heatcell{-67.05} & \heatcell{-43.31} & \heatcell{4.15} & \heatcell{-50.11} & \heatcell{41.08} & \heatcell{51.21} & \heatcell{54.93} & \heatcell{48.53} & \heatcell{51.80} & \heatcell{56.59} \\
\midrule
\multirow{2}{*}{Qwen-7B}
& Vanilla & \heatcell{-66.95} & \heatcell{-59.56} & \heatcell{-31.57} & \heatcell{-48.54} & \heatcell{36.76} & \heatcell{56.22} & \heatcell{63.00} & \heatcell{56.05} & \heatcell{60.63} & \heatcell{66.72} \\
& Persona & \heatcell{-64.97} & \heatcell{-56.66} & \heatcell{-8.86} & \heatcell{-50.20} & \heatcell{41.06} & \heatcell{55.61} & \heatcell{59.98} & \heatcell{53.49} & \heatcell{60.84} & \heatcell{65.24} \\
\bottomrule
\end{tabular}
\caption{Spearman rank correlation ($\times$100) 
between LLM attribute predictions and human ratings 
for small models, ordered from most negative to 
most positive.}
\label{tab:spearman-small}
\end{table*}

\begin{table}[t]
\centering
\small
\begin{tabular}{p{0.1cm}cccccc}
\toprule
& & {R²} & {F1} & {Acc.} & {Prec.} & {Rec.} \\
\midrule
\multirow{7}{*}{\rotatebox[origin=c]{90}{Llama-8B}}
& Zero-shot & -- & 49.05 & 74.50 & 51.86 & 46.54 \\
& Few-shot & -- & 50.19 & 47.88 & 33.55 & 99.54 \\
& Definition & -- & 59.86 & 74.40 & 51.04 & 72.38 \\
& Attr. aware & -- & 61.99 & 70.13 & 46.65 & 92.33 \\
& Attr. value & -- & 60.19 & 68.57 & 45.19 & 90.07 \\
& Vanilla & 62.55  & 56.34 & 80.97 & 71.36 & 46.54 \\
& Persona & 61.03  & 60.42 & 80.86 & 66.49 & 55.37 \\
\midrule
\multirow{7}{*}{\rotatebox[origin=c]{90}{Qwen-7B}}
& Zero-shot & -- & 58.08 & 63.61 & 41.72 & 95.55 \\
& Few-shot & -- & 54.38 & 56.69 & 37.65 & 97.84 \\
& Definition & -- & 57.81 & 63.59 & 41.63 & 94.58 \\
& Attr. aware & -- & 58.53 & 64.40 & 42.25 & 95.22 \\
& Attr. value & -- & 50.05 & 47.69 & 33.45 & 99.36 \\
& Vanilla & 60.53  & 56.11 & 80.37 & 68.39 & 47.56 \\
& Persona & 57.92 & 54.57 & 79.86 & 67.39 & 45.85 \\
\bottomrule
\end{tabular}
\caption{Hate speech score reconstruction and 
classification results ($\times$100) for small 
models. R² is reported as mean $\pm$ std over 
5 folds. Baselines are zero-shot direct prompting 
variants. Our pipeline consistently achieves 
superior accuracy and precision despite lower F1 
on some baseline variants.}
\label{tab:results-small}
\end{table}

Table~\ref{tab:results-small} presents the hate speech score reconstruction and classification results for the two smaller models. Our confidence-weighted Ridge pipeline achieves R² of up to 62.55 for Llama-3.1-8B and 60.53 for Qwen-2.5-7B, demonstrating that attribute-level predictions from smaller models can still recover a meaningful hate speech signal, though with a notable drop compared to their larger counterparts (70.57 and 68.85, respectively).

Compared to direct prompting baselines, the picture is more nuanced for smaller models than for large ones. While our pipeline consistently achieves superior accuracy and precision across both models, some baselines, particularly attribute-aware prompting for Llama-8B (F1: 61.99), achieve competitive or slightly higher F1 than our pipeline (56.34). This is largely driven by the same high recall/low precision pattern observed in large model baselines, where the model flags the vast majority of comments as hate speech. Our pipeline maintains substantially better precision (71.36 vs. 46.65 for Llama-8B), indicating more reliable, calibrated predictions. Notably, Llama-8B few-shot nearly collapses, achieving 99.54\% recall by predicting almost everything as hate speech, suggesting that few-shot prompting is unstable for smaller models on this task.

\section{Ablation Study}
\label{sec:ablation}

Table~\ref{tab:ablation} reports the ablation results for the hate speech score reconstruction pipeline across four formula variants, evaluated on the vanilla condition for both large models. The variants are:
\begin{itemize}
\item A: No Ridge, confidence + Spearman weighted sum
\item B: Ridge with confidence weighting (our method)
\item C: Ridge without confidence weighting
\item D: No Ridge, Spearman weighted sum only
\end{itemize}

The results reveal three clear findings. First, Ridge regression is essential; formulas A and D, which use weighted sums without Ridge, collapse to strongly negative R² values (as low as -13.24 for Llama and -10.15 for Qwen), confirming that a simple weighted aggregation of attribute predictions cannot recover the continuous hate speech score. Without Ridge, the inverted evaluative attributes directly subtract from the score in an uncontrolled way, producing unreliable predictions.

Second, confidence weighting improves performance; formula C, which uses Ridge without confidence weighting, achieves a lower R² than our method B (68.63 vs 70.57 for Llama-70B, 68.27 vs 68.85 for Qwen-72B). While the gain is modest, it is consistent across both models, supporting the intuition that a confident prediction on an attribute carries more signal than an uncertain one.

Third, Spearman pre-scaling is redundant; adding it on top of confidence weighting before Ridge produces identical results, confirming that Ridge absorbs the Spearman pre-scaling into its learned weights automatically. This simplifies the final pipeline to confidence weighting followed by Ridge regression.

\begin{table}[t]
\centering
\small
\begin{tabular}{llccc}
\toprule
& \textbf{Formula} & \textbf{R²} & \textbf{Pearson} & \textbf{Spearman} \\
\midrule
\multirow{4}{*}{\rotatebox[origin=c]{90}{Llama-70B}}
& A & -8.44 & 0.83 & 0.83 \\
& B & 70.57 & 0.84 & 0.83 \\
& C& 68.63 & 0.83 & 0.82 \\
& D & -13.24 & 0.82 & 0.81 \\
\midrule
\multirow{4}{*}{\rotatebox[origin=c]{90}{Qwen-72B}}
& A & -7.38 & 0.83 & 0.81 \\
& B & 68.85 & 0.83 & 0.82 \\
& C & 68.27 & 0.83 & 0.81 \\
& D & -10.15 & 0.82 & 0.81 \\
\bottomrule
\end{tabular}
\caption{Ablation study comparing formula variants 
for hate speech score reconstruction (vanilla 
condition, large models only). C = confidence 
weighting. R² reported 
as percentage ($\times$100).}
\label{tab:ablation}
\end{table}

\section{Qualitative Examples}
\label{sec:examples}
Table~\ref{tab:examples} presents three representative comments illustrating LLM attribute prediction behavior across the hate speech severity spectrum. The hate speech score threshold for binary classification is 0.5, following \citet{kennedy2020constructing}; comments with a score above this threshold are classified as hate speech. In the supportive trans community comment (ground truth: -8.34), Llama correctly rates behavioral attributes such as insult, violence, and genocide as zero, but incorrectly assigns high scores to sentiment and respect, a direct manifestation of the evaluative attribute inversion identified in Table~\ref{tab:spearman}. In the extreme hate speech example (ground truth: 6.30), Llama correctly captures behavioral dimensions but rates sentiment and respect as zero when humans rate them at 4, causing the Ridge pipeline to severely underestimate the final score (1.875 vs 6.30). The two intermediate examples further confirm that behavioral attributes are predicted more reliably than evaluative ones, and that Ridge predictions closely track ground truth when evaluative attributes are less central to the comment's hateful content.

\begin{table}[t]
\small
\begin{tabular}{lcc}
\toprule
& \textbf{Human} & \textbf{Llama} \\
\midrule
\multicolumn{3}{p{7cm}}{\textit{"Fellow trans girl here! Love and support confirmed!"}} \\
{Hate Score} & -8.34 & -4.83 \\
\midrule
sentiment   & 0.00 & 4.00 \\
respect     & 0.00 & 4.00 \\
status      & 2.00 & 4.00 \\
insult      & 0.00 & 0.00 \\
humiliate   & 0.00 & 0.00 \\
dehumanize  & 0.00 & 0.00 \\
violence    & 0.00 & 0.00 \\
genocide    & 0.00 & 0.00 \\
attack\_defend & 0.25 & 0.00 \\
\midrule
\multicolumn{3}{p{7cm}}{\textit{"They ain't got no new n***** coming out???"}} \\
{Hate Score} & -0.70 & -1.03 \\
\midrule
sentiment   & 3.25 & 1.00 \\
respect     & 3.25 & 1.00 \\
status      & 2.25 & 1.00 \\
insult      & 2.75 & 3.00 \\
humiliate   & 2.75 & 3.00 \\
dehumanize  & 2.25 & 3.00 \\
violence    & 1.25 & 0.00 \\
genocide    & 0.50 & 0.00 \\
attack\_defend & 2.25 & 3.00 \\
\midrule
\multicolumn{3}{p{7cm}}{\textit{"Dig burn PITS and treat them THE INVADING N**** ALL OVER EUROPE..."}} \\
{Hate Score} & 6.30 & 1.88 \\
\midrule
sentiment   & 4.00 & 0.00 \\
respect     & 4.00 & 0.00 \\
status      & 4.00 & 0.00 \\
insult      & 4.00 & 4.00 \\
humiliate   & 4.00 & 4.00 \\
dehumanize  & 4.00 & 4.00 \\
violence    & 4.00 & 4.00 \\
genocide    & 4.00 & 4.00 \\
attack\_defend & 4.00 & 4.00 \\
\bottomrule
\end{tabular}
\caption{Representative examples showing human vs 
Llama-70B attribute predictions and Ridge hate 
speech score. Evaluative attributes (sentiment, 
respect, status) show systematic misalignment, 
while behavioral attributes are predicted more 
reliably.}
\label{tab:examples}
\end{table}

\section{Use of Artifacts}
We use the Measuring Hate Speech (MHS) corpus, which is publicly available on HuggingFace under a Creative Commons Attribution 4.0 International (CC BY 4.0) license. This license permits use for research purposes. We use Meta-Llama-3.1-70B-Instruct and Meta-Llama-3.1-8B-Instruct, released under the Llama 3.1 Community License, which permits research and commercial use. We use Qwen2.5-72B-Instruct and Qwen2.5-7B-Instruct, released under the Apache 2.0 License, which permits research and commercial use. We use vLLM, released under the Apache 2.0 License, for model serving. Our experimental code will be made publicly available upon publication. 
The experiments were carried out on machines with NVIDIA GeForce GTX 1080 Ti GPUs. During evaluation, models were used in inference mode without any parameter updates. 

\section{AI Usage Statement}
We used AI assistance in several stages of this work. Claude (Anthropic) was used as a writing assistant to help draft, revise, and improve the clarity of the paper text. All AI-generated text was reviewed, edited, and verified by the authors. All experimental results, analyses, and scientific claims are the sole responsibility of the authors.

\end{document}